\documentclass[10pt,twocolumn,letterpaper]{article}

\usepackage{cvpr}              

\usepackage{parskip}
\usepackage{amssymb}
\usepackage{booktabs}
\usepackage{algorithm}
\usepackage{algorithmic}
\usepackage{graphicx}
\usepackage{amsmath}
\usepackage{amssymb}
\usepackage{booktabs}
\usepackage{multirow}
\usepackage{booktabs}
\usepackage{array}
\usepackage{threeparttable}
\usepackage{setspace}  
\usepackage{color}
\usepackage{subfloat}
\usepackage{caption}
\usepackage{subcaption}
\usepackage{newfloat}
\usepackage{listings}
\usepackage[pagebackref,breaklinks,colorlinks]{hyperref}

\usepackage[capitalize]{cleveref}
\crefname{section}{Sec.}{Secs.}
\Crefname{section}{Section}{Sections}
\Crefname{table}{Table}{Tables}
\crefname{table}{Tab.}{Tabs.}

\def\cvprPaperID{*****} 
\def\confName{CVPR}
\def\confYear{2023}

\title{General Point Model Pretraining with Autoencoding and Autoregressive}
	
\author{Zhe Li \textsuperscript{1, ${\dagger}$}, Zhangyang Gao \textsuperscript{3, ${\dagger}$}, Cheng Tan \textsuperscript{3, ${\dagger}$},  Stan Z. Li \textsuperscript{3, ${*}$}, Laurence T. Yang \textsuperscript{1,2, ${*}$}\\
\textsuperscript{1} Huazhong University of Science and Technology \\
\textsuperscript{2} Hainan University \\
\textsuperscript{3} AI Lab, Research Center for Industries of the Future, Westlake University \\
\thanks{$^{\dagger}$Equal Contribution, $^{*}$Corresponding Author.}
{\tt\small keycharon0122@gmail.com,  \{gaozhangyang,tancheng,stan.zq.li\}@westlake.edu.cn,\
ltyang@ieee.org}
}

\makeatletter
\def\thanks#1{\protected@xdef\@thanks{\@thanks
        \protect\footnotetext{#1}}}
\makeatother

\begin{document}
\maketitle

\begin{abstract}
	The pre-training architectures of large language models encompass various types, including autoencoding models, autoregressive models, and encoder-decoder models. We posit that any modality can potentially benefit from a large language model, as long as it undergoes vector quantization to become discrete tokens. Inspired by GLM, we propose a General Point Model (GPM) which seamlessly integrates autoencoding and autoregressive tasks in point cloud transformer. This model is versatile, allowing fine-tuning for downstream point cloud representation tasks, as well as unconditional and conditional generation tasks. GPM enhances masked prediction in autoencoding through various forms of mask padding tasks, leading to improved performance in point cloud understanding. Additionally, GPM demonstrates highly competitive results in unconditional point cloud generation tasks, even exhibiting the potential for conditional generation tasks by modifying the input's conditional information. Compared to models like Point-BERT, MaskPoint and PointMAE, our GPM achieves superior performance in point cloud understanding tasks. Furthermore, the integration of autoregressive and autoencoding within the same transformer underscores its versatility across different downstream tasks.
\end{abstract}

\section{Introduction}
In recent years, the natural language processing (NLP) \cite{brown2020language, kenton2019bert, joshi2020spanbert, liu2019roberta, radford2019language} and computer vision (CV) \cite{bao2021beit, chen2020generative, dosovitskiy2020image, touvron2021training, xie2021self, zhu2020deformable} realms have witnessed a proliferation of transformer-based pretrained models. Their primary advantage lies in the inclusion of massive parameters and data in the training process, overcoming inductive biases introduced by traditional Convolutional Neural Networks \cite{krizhevsky2012imagenet}. Point clouds serve as fundamental data structures in fields like autonomous driving and robotics, thus emphasizing the escalating significance of tasks related to point cloud representation learning and generation. However, the realm of pretrained point cloud models based on transformers remains relatively scarce. Existing transformer-based point cloud models \cite{guo2021pct, zhao2021point} encounter inevitable inductive biases due to local feature aggregation \cite{guo2021pct}, neighbor embedding \cite{zhao2021point}, and the scarcity of annotated data \cite{dosovitskiy2020image}. These biases deviate from the mainstream of standard Transformers. Self-supervised models \cite{li2023vipmm, he2022masked, kenton2019bert, radford2019language, radford2018improving, gao2021simcse, yang2019xlnet} have emerged as the dominant methodology. They excel in learning high-quality representations across modalities without extensive labeled data, models like GPT \cite{radford2018improving} even dominate in text generation. In this end, there is a compelling need to design a transformer-based point cloud model that minimizes inductive biases, learns superior point cloud representations from limited data, and simultaneously possesses the capacities for conditional and unconditional point cloud generation.
\begin{figure}
	\centering 
	\includegraphics[scale=0.35]{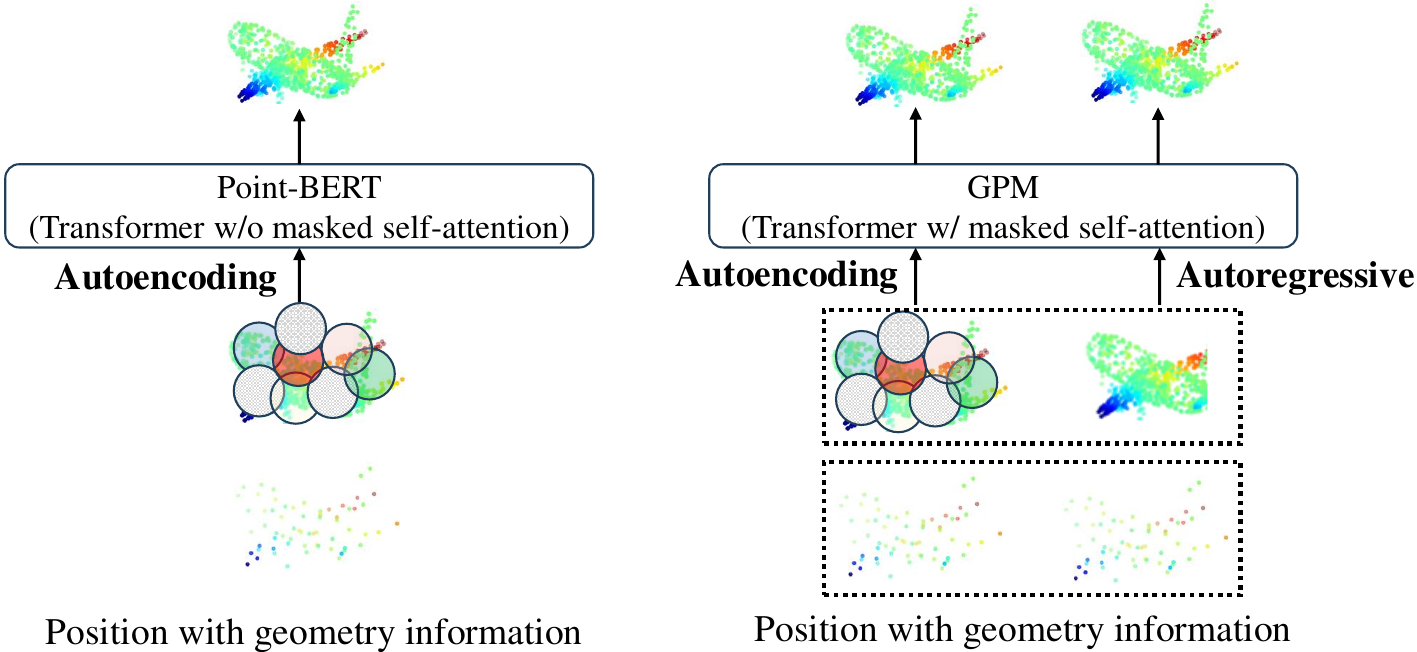} 
	\caption{Comparison of pre-training frameworks between GPM and Point-BERT \cite{yu2022point}. Point-BERT (left) conducts the mask prediction task in an autoencoding manner, while GPM (right) combines both autoencoding and autoregressive tasks, performing tasks for both mask prediction and mask generation separately.}  
	\label{motivation} 
\end{figure}
The transformer-based BERT \cite{kenton2019bert} has marked a significant milestone in the field of natural language processing (NLP). Its pre-training phase, involving masked prediction tasks, grants the model the capacity to learn language representations, showcasing the prowess of autoencoding language models. Given BERT's proficiency in learning language representations, the question arises: can we extend BERT's capabilities to the realm of point clouds? However, since point clouds lack a direct analog to the concept of 'words' in language, constructing a discrete vocabulary for point clouds, as done in language models, is infeasible. Point clouds are composed of individual points, and considering each point as a token would lead to a quadratic increase in computational cost as the number of tokens grows. Furthermore, each point in a point cloud contains limited semantic information, necessitating the maximization of geometric information utilization. Inspired by Point-BERT \cite{yu2022point}, we embarked on a journey to vector quantization, transforming point clouds into discrete tokens. Through the application of the farthest point sampling (FPS) \cite{qi2017pointnet++} algorithm, we segment the point cloud into multiple patches, each patch encapsulates plentiful geometric information, akin to a unit. By constructing a vocabulary specific to the point cloud domain, we obtain labels for each unit, discretizing the entire point cloud. This paves the way for the completion of the autoencoding model's masked prediction tasks in the point cloud domain.

Transformer-based GPT \cite{radford2018improving} has dominated the field of text generation, with its core being an autoregressive language model, predicting the next token based on preceding tokens. In recent years, researchers have progressively extended this paradigm into the realm of images \cite{chen2020generative, ramesh2021zero, ramesh2022hierarchical}, ushering in a new era of image generation. In this endeavor, we bring this paradigm to the domain of point clouds, aiming to empower it for autoregressive conditional and unconditional point cloud generation tasks. After discretization, each patch of the point cloud possesses distinct geometric information, enabling it to excel in autoregressive tasks. 

While these pre-training frameworks can be adapted to the point cloud domain, they lack the flexibility to satisfy all point cloud tasks. Therefore, we aim to integrate these two pre-training frameworks from the NLP domain and apply them to the point cloud domain, depicted in Figure \ref{motivation}.

In this work, we introduce a novel approach termed \textbf{G}eneral \textbf{P}oint \textbf{M}odel (GPM) with autoencoding and autoregressive, which integrates point cloud representation learning and point cloud generation (both conditional and unconditional) within the same transformer framework. Similar to Point-BERT, we employ a point cloud tokenizer designed through dVAE-based point cloud reconstruction, where the point cloud can be discretely labeled based on the learned vocabulary. Additionally, we utilize the Masked Point Modeling (MPM) task. This involves partitioning the point cloud into distinct patches and randomly masking some adjacent patches. This aims to predict the masked portion, with the goal of learning geometric information between neighboring blocks and capturing meaningful geometric features for understanding the point cloud. 

We divide the input into two segments, where the first accomplishes the masked point prediction task in an autoencoding manner, while another performs the masked generation task in an autoregressive manner. By designing specific attention mask matrix, both tasks are conducted within the same transformer. Simultaneously executing these two mask reconstruction tasks enables the model to more efficiently capture the geometric features of the point cloud and gain a more comprehensive understanding of it compared to Point-BERT. Moreover, with the conditioning information from the first part, the autoregressive generation in another is notably enhanced, highlighting the model's potential in both unconditional and conditional point cloud generation tasks (such as text/image-conditioned point clouds).

\section{Related Work}
\subsection{Self-supervised Learning}
Self-supervised learning (SSL) has garnered significant attention owing to the substantial labor required for acquiring annotated data in large quantities. SSL enables models to learn feature representations from unlabeled data, adapting to downstream tasks. The core of SSL lies the design of proxy tasks to replace traditional classification tasks, thereby learning feature representations. ELMo \cite{sarzynska2021detecting} employs bidirectional LSTMs \cite{hochreiter1997long} and generates subsequent words from left to right given the representation of preceding content. ViP \cite{li2023vipmm} employs a dynamically updated momentum encoder for contrastive learning and designs a text swapping task to enhance the model's sentence representation capability. In the computer vision domain, Image-GPT \cite{chen2020generative} trains a sequence transformer to autoregressively predict pixels, showing promising representation learning capabilities without incorporating specific knowledge about the 2D input structure after pretraining. 
Furthermore, self-supervised learning (SSL) in the field of point clouds has garnered significant attention. ACT \cite{dong2022autoencoders} employs cross-modal autoencoders as teacher models to acquire knowledge from other modalities. Most relevant to our work are generative approaches \cite{li2018so, achlioptas2018learning, sauder2019self, min2022voxel, yu2022point, zhang2022point}. Point-MAE \cite{pang2022masked} extends MAE by randomly masking point patches and reconstructing the masked regions. Point-M2AE \cite{zhang2022point} further utilizes a hierarchical transformer architecture and designs corresponding masking strategies. However, mask-based point modeling methods still face the issue of shape leakage, limiting their effective generalization to downstream tasks. In this paper, we leverage autoregressive pretraining on point clouds and address the unique challenges associated with point cloud attributes. Our concise design avoids position information leakage, thereby enhancing generalization capabilities.
\subsection{General Language Model}
In the field of natural language processing, GLM \cite{du2022glm} integrates both autoregressive and autoencoding methods. This enables it to perform tasks related to sentence representation as well as conditioned and unconditioned generation through fine-tuning. To the best of our knowledge, we are the first to integrate autoregressive and autoencoding techniques on a single transformer in the domain of point clouds. In downstream tasks, we have the potential to fine-tune point cloud representations and perform unconditioned generation even in zero-shot scenarios, with the possibility of fine-tuning towards conditioned generation.
\begin{figure*}
	\centering 
	\includegraphics[scale=0.5]{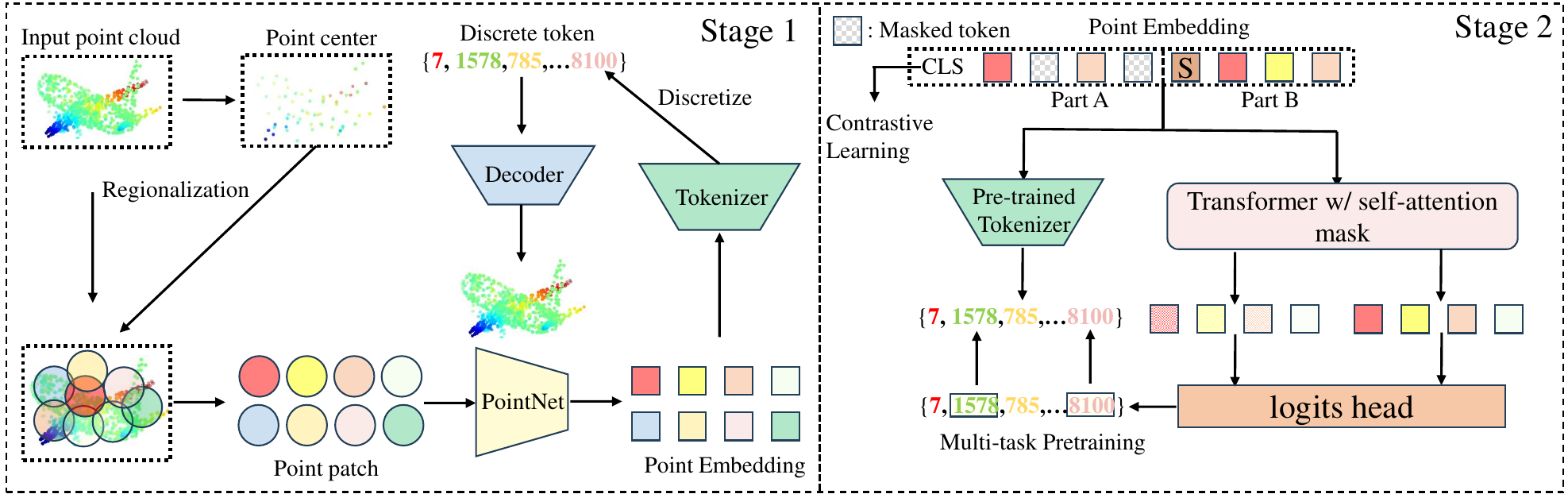} 
	\caption{The framework of GPM. We divide the entire pipeline into two stages: 1) The input point cloud is divided into several sub-clouds. Then, we utilize Mini-PointNet \cite{qi2017pointnet} to obtain a sequence of point embeddings. These embeddings are transformed into a sequence of discrete point tokens using a pre-trained tokenizer; 2) During pre-training, for PartA, certain portions of the masked embeddings are masked and replaced with [M]. They are predicted in an autoencoding manner. For PartB, masks for PartA are generated in an autoregressive manner.}  
	\label{framework} 
\end{figure*}
\section{Methods}
In this study, we endeavor to seamlessly integrate BERT-style and GPT-style pre-trained strategies, extending their applicability to point cloud transformers. Following the Point-BERT\cite{yu2022point}, our training process unfolds in two distinctive stages. Firstly, we embark on training a specialized Tokenizer, a critical step towards acquiring discrete point labels for each input point cloud. Concurrently, a dedicated dVAE is employed to masterfully reconstruct the discrete point cloud. The subsequent stage involves the splicing of two sequences of point cloud embeddings, partA for autoencoding and partB for autoregressive. The overall idea of our approach is illustrated in Figure \ref{framework}.
\subsection{Stage 1:Discrete Varitional Autoencoder Pre-training}
\paragraph{Point Cloud Partitioning.}
We posit that discrete tokens derived from points encapsulate crucial geometric information, allowing for a discrete representation of any point cloud. However, a naive strategy of assigning one token to each point in a point-wise reconstruction task poses a formidable computational challenge. This arises from the quadratic complexity of self-attention within transformers. Building upon the foundations laid by Point-BERT \cite{yu2022point} and ViT\cite{dosovitskiy2020image}, we adopt a strategy of partitioning each point cloud into distinct patches, each serving as a single token. Specifically, for an input point cloud $p \in \mathbb{R}^{\mathcal N \times 3}$, we initiate the process by extracting $m$ center points $\mathcal C_u \in \mathbb{R}^{m \times 3}$ from the point cloud $p$ through farthest point sampling (FPS). Subsequently, we employ the k-nearest neighbor (kNN) algorithm to identify the $k$ nearest neighboring points for every center point. This forms $m$ localized patches or sub-clouds denoted as $\mathcal P_u \in \mathbb{R}^{m \times k \times 3}$. To ensure these patches are free of bias, we normalize them by subtracting their respective center coordinates. This operation effectively disentangles the structural patterns from the spatial coordinates inherent to each local patch.  
\paragraph{Point Cloud Embedding.}
Follwing Point-BERT \cite{yu2022point}, we employ a mini-PointNet \cite{qi2017pointnet} to embed the point patches as a sequence of point embedding $\{h_i\}_{i=1}^{m}$. For the further vector quantization,  we pre-define a learnable codebook $\mathcal V := \{(s, z_s)\}_{s \in S}$, where $d_z$ is the dimension of codes, $S$ is the size of codebook and $s$ is the index of embedding in $\mathcal V$. We adopt DGCNN \cite{wang2019dynamic} as tokenizer $\mathcal Q: h_i \mapsto z_i$, which maps $\{h_i\}_{i=1}^{m}$ into $\{z_i\}_{i=1}^{m}$ in $\mathcal V$. However, given the non-differentiable nature of the discrete tokens, applying reparameterization gradients for dVAE training becomes unfeasible. As suggested in \cite{ramesh2021zero}, we resort to the Gumbel-softmax relaxation technique \cite{jang2016categorical}, coupled with a uniform prior, as a workaround during the dVAE training process.
\paragraph{Point Cloud Reconstruction.}
Given $\{z_i\}_{i=1}^{m}$ as input, the decoder $\mathcal D(\cdot)$ is tasked with the reconstruction of the entire point cloud. Inspired by \cite{yu2022point}, to effectively capture inter-point relationships and bolster the representational capacity of discrete point tokens across various local structures, we adopt the DGCNN \cite{wang2019dynamic}-FoldingNet \cite{yang2018foldingnet} architecture to reconstruct the whole point cloud.

The overarching reconstruction objective can be denoted as $\mathbb{E}_{z \sim \mathcal Q(z|h)}\log D(\hat{p}|z)$. Inspired by Point-BERT \cite{yu2022point}, the entire reconstruction process can be conceptualized as the maximization of the Evidence Lower Bound (ELB) for the log-likelihood, denoted as $\mathcal D(\hat{p}|p)$:
\begin{equation}\label{1}
	\begin{split}
		{\sum\limits}_{i=1}^{\mathcal N}\log D(\hat{p}_i|p_i) \geq &{\sum\limits}_{i=1}^{\mathcal N} (\mathbb{E}_{z_i} \sim \mathcal Q(z_i|p_i))[\log D(p_i|z_i)] \\
		&- \text{KL}[\mathcal Q(z|p_i), \mathcal D(z|\hat{p}_i)].
	\end{split}
\end{equation}
Simultaneously, we account for an intuitive reconstruction loss, leveraging the $l1$ Chamfer Distance \cite{fan2017point} to supervise prediction point cloud with the ground-truth point cloud:
\begin{equation}\label{1.5}
	\mathcal L_{\text{CD}}(\mathcal P,\mathcal G) = \frac{1}{\left| \mathcal P\right|}\sum_{p\in\mathcal P}\min\limits_{g \in \mathcal G}\Vert p-g \Vert+\frac{1}{\left| \mathcal G\right|}\sum_{g\in\mathcal G}\min\limits_{p \in \mathcal P}\Vert g-p \Vert,
\end{equation}
where $\mathcal P$ denotes the predicted point clouds and $\mathcal G$ is the ground truth point clouds. Additional, we follow \cite{ramesh2021zero}, optimize the KL-divergence $\mathcal L_{\text{KL}}$ between the generated point cloud distribution and a uniform distribution:
\begin{equation}\label{1.5}
	\mathcal L_{\text{dVAE}} = \mathcal L_{\text{CD}} + \mathcal L_{\text{KL}}.
\end{equation}
\subsection{Stage 2: GPM Pre-training}
To ensure GPM excel in both point cloud representation learning and point cloud generation task, we formulate a training framework that amalgamates autoencoding and autoregressive techniques motivated by \cite{du2022glm}.
\paragraph{Autoencoding Masked Sequence Generation.}
We employ the standard Transformers \cite{vaswani2017attention}, which encompass multi-headed self-attention layers and FFN blocks. Initially, we partition each input point cloud into $m$ local patches with designated center points. Subsequently, these local patches undergo discretization into a code sequence  $\{f_i\}_{i=1}^{m}$ via our pre-trained dVAE in stage 1. Formally, the input embeddings $\{e_i\}_{i=1}^{m}$ are constructed as a combination of point embeddings $\{f_i\}_{i=1}^{m}$ and positional embeddings $\{pos_i\}_{i=1}^{m}$. Adhering to language model \cite{kenton2019bert}, a class token $[\text{CLS}]$ is appended to the input sequences, rendering the transformer input sequence as $I = \{[\text{CLS}], e_1^0, e_2^0, ..., e_i^0\}$.

Drawing inspiration from \cite{yu2022point}, we initiate by selecting a central point $\mathcal C_i$ alongside its corresponding sub-cloud $\mathcal P_i$, integrate it with $m$ neighboring sub-clouds as a coherent local region. Within this region, we apply a masking operation to obscure all local patches, generating what we refer to as the "masked point cloud". Specifically, we substitute all the masked point embeddings with a universally learnable pre-defined mask embeddings $[\text{M}]$. We mark the masked position as $\mathcal M = \{1, ..., b\}$, and the final input embeddings $E^{\mathcal M}=\{e_i ,i \notin \mathcal M\} \cup \{[\text{M}]+pos_i, i\in \mathcal M\}$ are fed into transformer.

The objective of our Masked Point Modeling (MPM) task is to deduce the geometric arrangement of absent portions and reconstruct point tokens aligned with the masked positions using the available information. Geometric information refers to the relative positions and morphological attributes of points within a localized region. These crucial geometric details provide vital cues for understanding and processing the shape and structure of the local region, playing a pivotal role in delving deep into the intricacies of point cloud data. Formally, the pre-training objective can be expressed as maximizing the log-likelihood of the accurate point tokens $e_i$ conditioned on the masked input embeddings $E^{\mathcal M}$:
\begin{equation}\label{2}
	\text{max}\sum_{E\in W}\mathbb{E}\bigg[\sum_{i \in \mathcal M}\log P(e_i|E^{\mathcal M})\bigg],
\end{equation}
where $W$ is the set of all input embeddings. Similar to \cite{yu2022point}, we also adopt Point Patch Mixing and contrastive learning to help the model to better learn high-level semantics.
\paragraph{Autoregressive Sequence Generation.}

While Point-BERT models the relationship between unmasked and masked regions using the Masked Point Modeling (MPM) task, it does not adequately model the interactions within masked regions. Furthermore, our aim extends beyond acquiring high-quality point cloud representations; we also aspire to proficiently execute point cloud generation tasks. Hence, for the latter segment, PartB, we engage in autoregressive mask generation tasks to enhance the MPM task.

In the case of PartB, the mask operation is ignored. Our focus lies in acquiring the generated tokens from the masked positions of PartA. We take the initial $n-1$ tokens from PartA, appending the start token $[\text{S}]$ at its forefront to get the PartB $E^{\mathcal B}=\{[\text{S}], e_1, ..., e_{n-1}\}$. Following the autoregressive approach, the prediction of the next token relies on the preceding token:
\begin{equation}\label{3}
	\max\limits_{\theta}\mathbb{E}\bigg[\sum_{i=1}^n\log P_{\theta}(e_i|\text{PartA}, e_{<i},[\text{S}])\bigg].
\end{equation}

Although no masking operation is performed on PartB, since the mask positions of PartA are multiple consecutive sub-clouds instead of a single one, in PartB we are also equivalent to autoregressive generation of a continuous mask span. We extract the generated tokens from PartB, corresponding to the masked positions in PartA, and compare them with the discrete tokens produced by the Tokenizer:
\begin{equation}\label{4}
	\mathcal L_{AR} = \text{CrossEntropy}(e_{j, j \in \mathcal M}, {\text{argmax}\mathcal Q(h)}_{j,j \in \mathcal M}).
\end{equation}

In this way, it allows our model to acquire proficiency in an autoencoding bidirectional encoding scheme for PartA, and an autoregressive unidirectional encoder for PartB.

\subsection{Multi-task Pretraining}
The masking of 15\% of tokens in BERT is tailored for downstream natural language understanding tasks, whereas Point-BERT masks 25\% $\sim$ 45\% to acquire enhanced point cloud representations. Our GPM focus on both point cloud representation learning and point cloud generation, we concatenate PartA and PartB as input to the transformer, aiming to simultaneously perform both autoencoding and autoregressive tasks. Therefore, we need to establish specific attention masks to facilitate the interaction of information between these two segments.

\begin{figure}
	\centering 
	\includegraphics[scale=0.7]{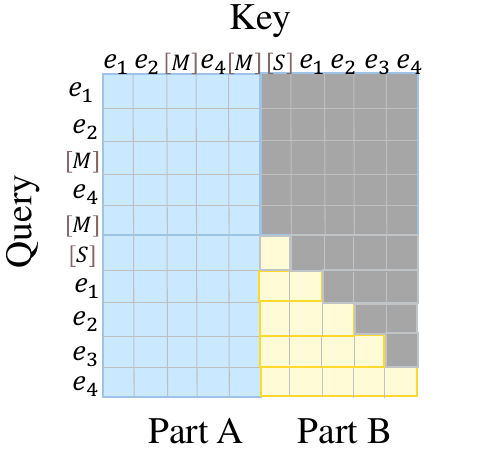} 
	\caption{Self-attention mask. The regions in grey are masked. Tokens in PartA can attend to themselves, but not to PartB. Tokens in PartB can attend to PartA and their antecedents in PartB.}  
	\label{attnmask} 
\end{figure}

Motivated by \cite{du2022glm}, the tokens in PartA can participate in MLM task in PartA and autoregressive generation in PartB. PartB tokens can not be observed by PartA, but those antecedents. PartA is treated as the conditioning for autoregressive generation in PartB, akin to prefix tuning \cite{li2021prefix} in NLP. The implement of GPM multi-task pretraining is depicted in Figure \ref{attnmask}.

The tokens output by the pretrained tokenizer $\mathcal Q(\cdot)$ are regarded as labels. PartA conducts a mask prediction task in an autoencoding manner, and the prediction loss between the predicted masked portion and the labels is computed as follows:
\begin{equation}\label{5}
	\mathcal L_{AE} = \text{CrossEntropy}(e_{j, j \in \mathcal M}, {\text{argmax}\mathcal Q(h)}_{j,j \in \mathcal M}).
\end{equation}
Although Equation \ref{4} and \ref{5} exhibit identical formal structures. However, one entails mask prediction through autoencoding, while the other involves mask generation in an autoregressive way, indicating a fundamental distinction between the two approaches.
\section{Experiments}
In this section, we commence by outlining the configurations of our pretraining scheme. Subsequently, we proceed to assess the proposed model through various downstream tasks, encompassing object classification, partial segmentation, few-shot learning, transfer learning (to validate the model's representation learning capabilities), and point cloud generation (to validate its generative capabilities). Additionally, we conduct ablation studies on GPM.
\subsection{Pre-training Setups and Implementation}
\paragraph{Pre-training Data.}
Following the dataset configuration similar to Point-BERT \cite{yu2022point}, we employ ShapeNet \cite{chang2015shapenet} as our pretraining dataset, encompassing over 50,000 unique 3D models spanning 55 common object categories. From each 3D model, we sample 1024 points, dividing them into 64 point patches (sub-clouds) and each sub-cloud contains 32 points. Utilizing a lightweight PointNet \cite{qi2017pointnet} with two MLP layers, we project each sub-cloud into a 64-point embedding, serving as input for both dVAE and Transformer \cite{vaswani2017attention}.
\paragraph{dVAE Pretraining Setups.}
The purpose of pretraining the dVAE is to acquire a high-quality tokenizer, enabling the reconstruction of discretized features back into the original point cloud to the fullest extent. Our dVAE is composed of a tokenizer and a decoder. To be specific, the tokenizer consists of a 4-layer DGCNN \cite{wang2019dynamic}, while the decoder encompasses a 4-layer DGCNN followed by a FoldingNet \cite{yang2018foldingnet}. Additionally, the FoldingLayer establishes a connection between a 2D grid and the input, ultimately generating a 3D point cloud. 

During the training of dVAE \cite{ramesh2021zero}, we set the number of words in the codebook $N$ to 8192. We employ the common $l1$-style Chamfer Distance loss in the reconstruction process. Due to the small numerical value of this $l1$ loss, the weight of the KL loss in Equation \ref{1} must be smaller than that in image tasks. In the initial 10,000 steps of training, the weight of the KL loss is kept at 0 and gradually raised to 0.1 in the subsequent 100,000 steps. Our learning rate is set to 0.0005, and follows a cosine annealing schedule spanning 60,000 steps. Consistent with prior works \cite{ramesh2021zero} and Point-BERT \cite{yu2022point}, we decay the temperature in the Gumble-softmax function from 1 to 0.0625 over 100,000 steps. The training of dVAE encompasses a total of 150,000 steps, with a batch size of 64. It's worth noting that our dVAE's network architecture closely mirrors that of Point-BERT. The detailed network architecture of our dVAE aligns with that in Point-BERT. 
\paragraph{GPM Pretraining Setups.}
In our experiments, we adhere to the standard Transformer architecture \cite{dosovitskiy2020image}, which consists of a stack of Transformer blocks \cite{vaswani2017attention}, each composed of multi-head self-attention layers and feed-forward networks. LayerNorm is applied in both layers. We set the depth of our Transformer to 12, feature dimension to 384, and the number of heads to 6. A random depth of 0.1 \cite{huang2016deep} is applied in our Transformer encoder. During GPM pretraining, the weights of the Tokenizer learned by dVAE are kept fixed. Input point embeddings from 25\% $\sim$ 45\% are randomly masked, and the model is trained to infer the expected point labels at these masked positions. Unlike Point-BERT \cite{yu2022point}, which does not have a generation task during pretraining, we incorporate both autoregressive and autoencoding tasks in our work. Thus, an attention mask is necessary, as illustrated in Figure \ref{attnmask}. For MoCo, we set the same configuration as \cite{yu2022point}.
\subsection{Downstream Tasks}
\subsubsection{Point Cloud Representation Evaluation}
\begin{table}[h]
	\setlength{\belowdisplayskip}{-1.5cm}
	\centering
	\resizebox{\linewidth}{!}{
		\begin{tabular}{cccc}
			\toprule
			{Methods} & {number of points} & {Acc}  \\
			\midrule
			\textit{\textbf{Supervised Learning}}\\
			PointNet\cite{qi2017pointnet} & 1k &89.2\\ 
			SO-Net\cite{li2018so} & 1k&92.5\\
			PointNet++ \cite{qi2017pointnet++}&1k &90.5\\
			PointCNN\cite{li2018pointcnn} & 1k &92.2\\
			DGCNN\cite{wang2019dynamic}&1k&92.9\\
			MVTN\cite{hamdi2021mvtn}& 1k & 93.8 \\
			RSCNN \cite{rao2020global}& 1k & 92.9\\
			GBNet\cite{qiu2021geometric} & 1k &93.8\\
			PointMLP\cite{ma2021rethinking} &1k&94.5\\
			DensePoint\cite{liu2019densepoint} &1k&92.8\\
			PointNeXt\cite{qian2022pointnext}& 1k&94.0\\
			P2P-RN101\cite{wang2022p2p}& 1k&93.1\\
			P2P-HorNet\cite{wang2022p2p}& 1k&94.0\\
			KPConv\cite{thomas2019kpconv} & $\sim$6.8k &92.9\\
			\midrule
			\textit{\textbf{Self-Supervised Representation Learning}}\\
			$[\text{T}]$PCT \cite{zhao2021point}&1k &93.2 \\
			$[\text{T}]$PointTransformer \cite{guo2021pct}& -& 93.7 \\
			$[\text{ST}]$NPCT \cite{guo2021pct}& 1k & 91.0 \\
			$[\text{ST}]$Transformer\cite{vaswani2017attention}& 1k & 91.4 \\
			$[\text{ST}]$Transformer-OcCo\cite{vaswani2017attention}& 1k & 92.1 \\
			$[\text{ST}]$Point-BERT\cite{yu2022point} & 1k &93.2\\
			$[\text{ST}]$MaskPoint\cite{liu2022masked} &1k &\textbf{93.8}\\
			$[\text{ST}]$Point-MAE\cite{pang2022masked} &1k&\textbf{93.8}\\
			\textbf{GPM}& 1k&\textbf{93.8}\\
			\midrule
			\textit{\textbf{Methods with cross-modal information and teacher models}}\\
			ACT\cite{dong2022autoencoders} &1k&93.7\\
			\midrule
			$[\text{ST}]$Transformer\cite{vaswani2017attention}& 4k & 91.2 \\
			$[\text{ST}]$Transformer-OcCo\cite{vaswani2017attention}& 4k & 92.2 \\
			$[\text{ST}]$Point-BERT\cite{yu2022point} & 4k &93.4\\
			\textbf{GPM} &4k&\textbf{93.9}\\
			\midrule
			$[\text{ST}]$Point-BERT\cite{yu2022point} & 8k &93.8\\
			$[\text{ST}]$MaskPoint\cite{liu2022masked} &8k &-\\
			$[\text{ST}]$Point-MAE\cite{pang2022masked} &8k&94.0\\
			\textbf{GPM} & 8k&\textbf{94.1}\\
			\bottomrule
	\end{tabular}}
	\caption{Classification results on ModelNet40. All results are expressed as percentages accuracy.}
	\label{Table 1}
\end{table}
\paragraph{Object classification on ModelNet40.}
We evaluate our pre-trained model on the ModelNet40 dataset \cite{wu20153d}, comprising 12,311 clean 3D CAD models spanning 40 categories. Following the Point-BERT \cite{yu2022point} setups, we employ a two-layer MLP with a dropout rate of 0.5 as the classification head for the task. We optimize the model using AdamW with a weight decay of 0.05 and a learning rate of 0.0005, while employing a batch size of 32 and a cosine annealing schedule. We conduct comparisons with various Transformer-based models with identical Point-BERT settings, denoting [ST] for the standard Transformer architecture and [T] for Transformer models with specific designs or inductive biases. The results in Table \ref{Table 1} demonstrate that our model not only outperforms Point-BERT in classification metrics on this dataset but also exhibits autoregressive generation capabilities.
\paragraph{Object classification on ScanObjectNN.}
The practical applicability of our model on real datasets serves as a crucial metric. Therefore, the pre-trained models are transferred to the ScanObjectNN dataset \cite{uy2019revisiting}, which comprises approximately 15,000 objects extracted from real indoor scans, encompassing 2902 point clouds from 15 categories. This dataset presents a greater challenge as it involves sampling from real-world scans with backgrounds and occlusions. We conducted experiments on three main variants, namely OBJ-BG, OBJ-ONLY, and PB-T50-RS, in line with prior work. The experimental results are summarized in Table \ref{Table 2}. It is observed that GPM exhibits an improvement of approximately 2.77\%, 1.88\%, and 1.73\% over the regular Point-BERT across the three variants.
\begin{table}[h]
	\setlength{\belowdisplayskip}{-1.5cm}
	\centering
	\resizebox{\linewidth}{!}{
		\begin{tabular}{ccccc}
			\toprule
			{Methods} & {OBJ-BG} & {OBJ-ONLY} &{PB-T50-RS}  \\
			\midrule
			\textit{\textbf{Supervised Learning}}\\
			PointNet\cite{qi2017pointnet} & 73.3 &79.2 & 68.0\\ 
			SpiderCNN\cite{xu2018spidercnn} & 77.1 &79.5 & 73.7\\
			PointNet++ \cite{qi2017pointnet++}& 82.3 &84.3 & 77.9\\
			PointCNN\cite{li2018pointcnn} & 86.1 &85.5 & 78.5\\
			DGCNN\cite{wang2019dynamic}&82.8&86.2&78.1\\
			MVTN\cite{hamdi2021mvtn}& 92.6 & 92.3 & 82.8\\
			BGA-DGCNN\cite{uy2019revisiting} & - &- & 79.7\\
			BGA-PN++ \cite{uy2019revisiting}& - &- & 80.2\\
			GBNet\cite{qiu2021geometric} &- &-&81.0\\
			PointMLP\cite{ma2021rethinking} &-&-&85.4\\
			PointNeXt\cite{qian2022pointnext}&-&-&87.7\\
			P2P-RN101\cite{wang2022p2p}&-&-&87.4\\
			P2P-HorNet\cite{wang2022p2p}&-&-&89.3\\
			\midrule
			\textit{\textbf{Self-Supervised Representation Learning}}\\
			Transformer\cite{vaswani2017attention}& 79.86 & 80.55 & 77.24 \\
			Transformer-OcCo\cite{vaswani2017attention}& 84.85 & 85.54 & 78.79 \\
			Point-BERT\cite{yu2022point} & 87.43 &88.12 & 83.07\\
			MaskPoint\cite{liu2022masked} &89.3 &88.1 &84.3\\
			Point-MAE\cite{pang2022masked} &90.0&88.2&\textbf{85.2}\\
			\midrule
			\textbf{GPM}  & \textbf{90.2}&\textbf{90.0}&84.8\\
			\bottomrule
	\end{tabular}}
	\caption{Classification results on ScanObjectNN. All results are expressed as percentages accuracy.}
	\label{Table 2}
\end{table}
\begin{figure*}
	\centering 
	\includegraphics[scale=0.41]{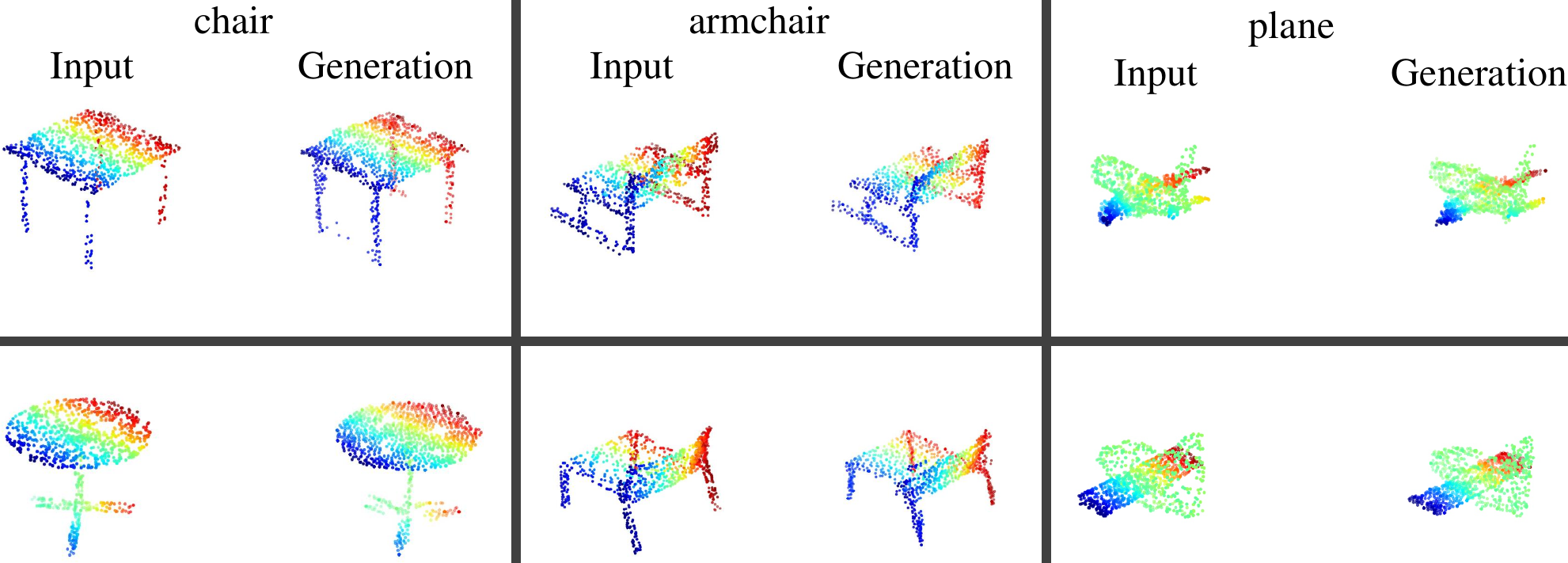} 
	\caption{Results on point cloud generation autoregressively. It is evident that we have largely reconstructed the original point cloud.}  
	\label{generation} 
\end{figure*}
\paragraph{Few-shot learning.}
\begin{table}[h]
		\centering
		\resizebox{\linewidth}{!}{
			\begin{tabular}{ccccc}
				\toprule
				\multirow{3}*{Methods} & \multicolumn{2}{c}{5-way} & \multicolumn{2}{c}{10-way}\\
				\cline{2-5}\\
				& 10-shot&20-shot&10-shot&20-shot\\
				\midrule
				\textit{\textbf{Supervised Learning}}\\
				DGCNN-rand \cite{wang2019dynamic}&31.6 $\pm$ 2.8&40.8  $\pm$ 4.6&19.9  $\pm$ 2.1 & 16.9 $\pm$ 1.5\\
				OcCo \cite{wang2021unsupervised}&90.6  $\pm$ 2.8&92.5 $\pm$ 1.9&82.9  $\pm$ 1.3 & 86.5  $\pm$ 2.2\\
				\midrule
				\textit{\textbf{Self-Supervised Representation Learning}}\\
				Point-BERT \cite{yu2022point}&94.6 $\pm$ 3.1&96.3  $\pm$ 2.7&91.0  $\pm$ 5.4 & 92.7  $\pm$ 5.1\\	
				MaskPoint \cite{liu2022masked}&95.0  $\pm$ 3.7&97.2  $\pm$ 1.7&91.4  $\pm$4.0 & 93.4  $\pm$ 3.5\\
				Piont-MAE \cite{pang2022masked}&96.3  $\pm$ 2.5&97.8 $\pm$ 1.8&92.6  $\pm$ 4.1 & 95.0  $\pm$ 3.0\\
				Piont-M2AE \cite{zhang2022point}&96.8  $\pm$ 1.8&98.3 $\pm$ 1.4&92.3  $\pm$ 4.5 & 95.0  $\pm$ 3.0\\
				\midrule
				\textbf{GPM} & \textbf{97.2 $\pm$5.3}&\textbf{98.7 $\pm$3.8}&\textbf{92.9 $\pm$4.7}&\textbf{95.0 $\pm$6.7}\\
				\bottomrule
		\end{tabular}}
		\caption{The results of few-shot classification on the ModelNet40 dataset. For each experimental setting, we conduct 10 independent experiments and report the average accuracy (\%) along with its standard deviation.}
		\label{Table 3}
	\end{table}
	To demonstrate the capability to acquire knowledge for new tasks under the constraints of limited training data, we evaluate our model under the setting of few-shot learning, following the methodology of previous work \cite{sun2017revisiting, yu2022point}. In the typical '$W$-way $S$-shot' setup, we initially randomly select $W$ classes and then sample ($S$+20) objects for each class [42]. The model is trained on $W \times S$ samples (support set) and evaluated on the remaining 20$W$ samples (query set). We conduct 10 independent experiments for each setting and report the average performance and standard deviation across the 10 runs. As shown in Table \ref{Table 3}, our approach outperforms other methods in all tests, achieving absolute improvements of 0.4\%, 0.4\%, and 0.6\% over Point-BERT.
	
	\paragraph{Part segmentation.}
	We evaluate the representation learning capability of our approach on the ShapeNetPart dataset \cite{yi2016scalable}, aiming to predict more fine-grained class labels for each point. This dataset consists of 16 categories and comprises 16,881 objects. The point cloud is downsampled to 2048 points, and the segmentation head \cite{pang2022masked} connects features $\mathcal F^4, \mathcal F^8, \mathcal F^{12}$ extracted from the 4-$th$, 6-$th$, and 12-$th$ layers of the transformer blocks. Subsequently, average pooling, max pooling, and upsampling are employed to generate features for each point, followed by label prediction using MLP. The experimental results shown in Table \ref{Table 4} demonstrate the superior performance of our GPM compared to all other methods.
	\begin{table}[h]
			\centering
			\resizebox{\linewidth}{!}{
				\begin{tabular}{c|cc}
					\toprule
					{Methods} & Cls.mIoU & Inst.mIoU \\
					\midrule
					\textit{\textbf{Supervised Learning}}\\
					PointNet\cite{qi2017pointnet} & 80.39 &83.7\\ 
					PointNet++ \cite{qi2017pointnet++}& 81.35 &85.1\\
					DGCNN\cite{wang2019dynamic}&82.33&85.2\\
					PointMLP \cite{ma2021rethinking}& 84.6 & 86.1\\
					\midrule
					\textit{\textbf{Self-Supervised Representation Learning}}\\
					Transformer\cite{vaswani2017attention}& 83.42 & 85.1 \\
					Transformer-OcCo\cite{vaswani2017attention}& 83.42 & 85.1 \\
					PointContrast \cite{xie2020pointcontrast} & -&85.1\\
					CrossPoint \cite{afham2022crosspoint}&-&85.5\\
					Point-BERT \cite{yu2022point} &84.11 &85.6\\
					Point-MAE \cite{pang2022masked}&-&\textbf{86.1}\\
					\midrule
					GPM & \textbf{84.20} & 85.8\\
					\bottomrule
			\end{tabular}}
			\caption{Part segmentation results on the ShapeNetPart dataset. We report the average intersection mIoU over the union of all classes (Cls.) and instances (Inst.).}
			\label{Table 4}
		\end{table}
		
		\subsubsection{Point Cloud Generation Evaluation}
		In the pre-training phase, we perform unconditional autoregressive generation on tokens in PartB, endowing the model with the capability of point cloud unconditional generation. Therefore, in downstream tasks, without fine-tuning, we conduct an autoregressive point cloud generation task on the ShapeNet dataset. Existing models do not possess both point cloud generation and point cloud classification tasks. Hence, our model is the first known point cloud transformer that integrates these two tasks into one. The generation results are depicted in the Figure \ref{generation}.
		\begin{figure*}[hbt!]
			\centering
			\vspace{-0.5cm} 
			\begin{subfigure}[b]{0.45\textwidth}
				\centering
				\includegraphics[width=\textwidth]{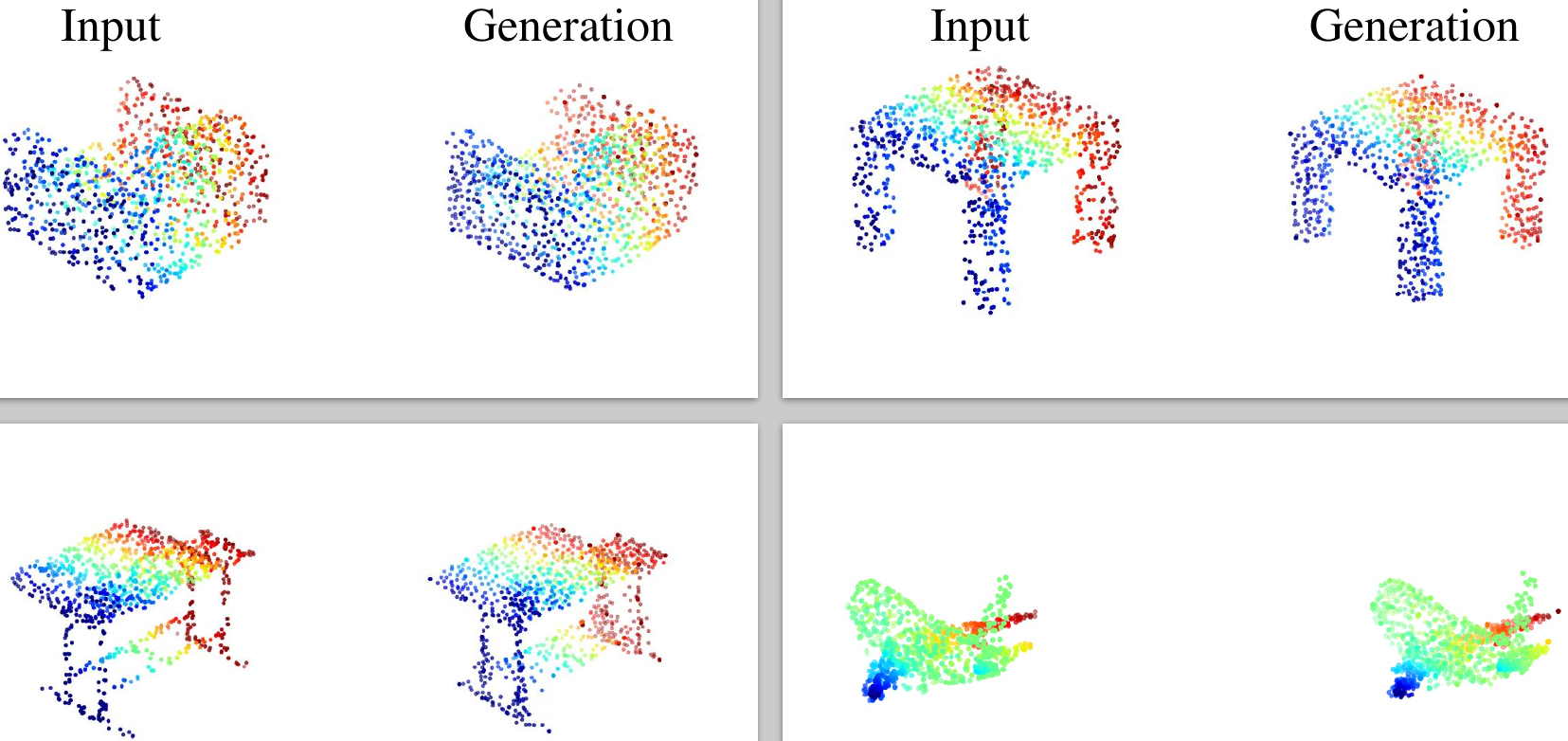}
				\caption{Point cloud generation results of dVAE.}
				\label{dvae}
			\end{subfigure}%
			\hfill
			\begin{subfigure}[b]{0.45\textwidth}
				\centering
				\includegraphics[width=\textwidth]{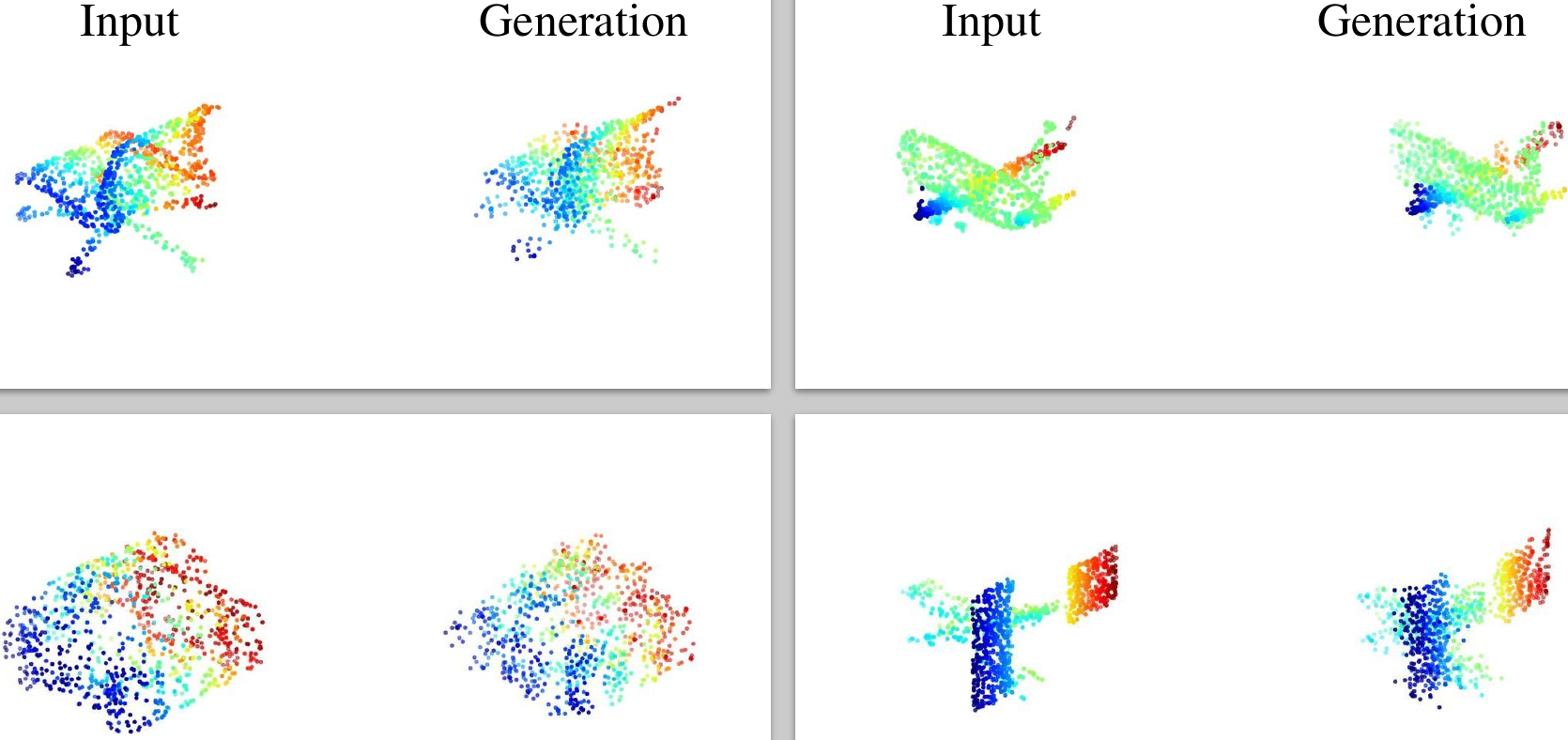}
				\caption{Point cloud generation results of Point-BERT.}
				\label{pointbert}
			\end{subfigure}%
			\caption{The reconstruction results visualization on dVAE and Point-BERT. }
			\label{vis}
		\end{figure*}
		\subsection{Ablation Study}
		To validate the fundamental design of the GPM model, we conducted a comprehensive set of ablation studies. In order to assess the impact of autoregressive mask generation on the autoencoding mask prediction task, we present results comparing fine-tuning scenarios on the ModelNet40 dataset: one with $\textbf{\textit{autoencoding}}$ only, and the other with a combination of $\textbf{\textit{autoencoding} + \textit{autoregreesive}}$. The results are summarized in the Table \ref{Table 5}. Furthermore, to demonstrate that the order of autoencoding and autoregressive tasks in the sequence does not affect point cloud understanding, we conducted additional experiments, with results also presented in the Table \ref{Table 6}.
		\begin{table}[h]\tiny
				\centering
				\resizebox{\linewidth}{!}{
					\begin{tabular}{c|c|c}
						\toprule
						{Tasks} & npoints&Acc \\
						\midrule
						autoencoding only & 1k&92.9\\
						\midrule
						autoencoding 
						+ autoregressive & 1k&\textbf{93.8}\\
						\bottomrule
				\end{tabular}}
				\caption{The comparison of results between $\textbf{\textit{autoencoding}}$ and $\textbf{\textit{autoencoding + autoregressive}}$ on the ModelNet dataset.}
				\label{Table 5}
			\end{table}
			\begin{table}[h]\tiny
					\centering
					\resizebox{\linewidth}{!}{
						\begin{tabular}{c|c|c}
							\toprule
							{Order} & npoints&Acc \\
							\midrule
							autoregressive-autoencoding & 1k&93.6\\
							\midrule
							autoencoding-autoregressive & 1k&\textbf{93.8}\\
							\bottomrule
					\end{tabular}}
					\caption{Results on different orders of autoregressive and autoencoding on ModelNet dataset. It can be observed that the order of autoencoding and autoregressive does not have a significant impact on point cloud understanding tasks.}
					\label{Table 6}
				\end{table}
				\subsection{Analysis}
				\paragraph{Compare with Point-BERT.}
				Motivated by BERT, Point-BERT leverages MPM as a pretext task for training. Due to the assumption of independence between MLM and MPM, Point-BERT fails to capture interdependencies among masked tokens. In our approach, the autoregressive generation task of masked tokens in PartB enhances the information exchange between them. This leads to improved performance in downstream tasks in the realm of representation learning. Simultaneously, tokens in PartA can attend to task in PartB, making PartA a generation condition for PartB, resulting in enhanced generation performance.
				\paragraph{Compare with PointGPT.}
				PointGPT \cite{chen2023pointgpt} adopts a two-stage Transformer architecture for learning representations of point clouds, comprising a feature extractor and a generator. While PointGPT exhibits decent generation capabilities, it doesn't perform autoregressive generation. Instead, it first extracts features from the point cloud using the feature extractor and then directly generates using the obtained features in the generator. During training, it directly employs Chamfer Distance \cite{fan2017point} between generated point clouds and ground truth points as the optimization objective, rather than operating on tokens. This is because it doesn't discretize points embeddings, thus failing to achieve genuine autoregressive generation tasks. Moreover, it doesn't integrate the two tasks for joint training in a single transformer; instead, it trains them separately on two different transformers.
				\subsection{Visualization}
				We visualize the generation results of dVAE and Point-BERT, as shown in the Figure \ref{dvae} and Figure \ref{pointbert}. Although Point-BERT is capable of masked tokens generation through autoencoding, our approach allows for autoregressive \textbf{conditional} generation tasks, such as text-conditioned point clouds and image-conditioned point clouds. As observed in Figure \ref{pointbert}, the generation results of Point-BERT exhibit some scattered points at the edges of the overall structure, appearing relatively sparse and disorganized. The generated outcome is not as satisfactory as that of GPM.
				
				\section{Conclusion}
				Our study introduces an innovative approach in the field of point clouds, integrating both autoencoding and autoregressive tasks within a unified transformer framework to enhance the model's performance in representation learning and generation tasks. By introducing the autoregressive generation task, we effectively enhance the information interaction among masked tokens during the training phase, enabling the model to excel in downstream tasks. Additionally, the autoencoding task effectively serves as a condition for the generation task, endowing the model with the potential for conditional generation, such as text-conditioned and image-conditioned point cloud generation. 
				{\small
					\bibliographystyle{ieee_fullname}
					\bibliography{egbib}
				}
				
			\end{document}